\documentclass[conference]{IEEEtran}
\IEEEoverridecommandlockouts
\usepackage{cite}
\usepackage{amsmath,amssymb,amsfonts}
\usepackage{algorithmic}
\usepackage{graphicx}
\usepackage{textcomp}
\usepackage{xcolor}
\usepackage{graphics}
\graphicspath{./pic/}
\usepackage{hyperref}
\usepackage{float}
\usepackage{subfigure}
\usepackage{pythonhighlight}
\usepackage[ruled,noend]{algorithm2e}
\usepackage[normalem]{ulem}
\usepackage{pifont}
\usepackage{bold-extra}

\begin{document}

\title{FTuner: A Fast Dynamic Shape Tensors Program Auto-Tuner for Deep Learning Compilers\\}

\author{\IEEEauthorblockN{1\textsuperscript{st} Pengyu Mu}
\IEEEauthorblockA{\textit{School of Computer Science and Engineering} \\
\textit{Beihang University}\\
Beijing, China \\
muphy@buaa.edu.cn}
\and
\IEEEauthorblockN{2\textsuperscript{th} Linquan Wei}
\IEEEauthorblockA{\textit{School of Computer Science and Engineering} \\
\textit{Beihang University}\\
Beijing, China}
\and
\IEEEauthorblockN{3\textsuperscript{rd} Rui Wang}
\IEEEauthorblockA{\textit{School of Computer Science and Engineering} \\
\textit{Beihang University}\\
Beijing, China}
\and
\IEEEauthorblockN{4\textsuperscript{nd} Yi Liu}
\IEEEauthorblockA{\textit{School of Computer Science and Engineering} \\
\textit{Beihang University}\\
Beijing, China}
}

\maketitle

\begin{abstract}
Many artificial intelligence models process input data of different lengths and resolutions, making the shape of the tensors dynamic. The performance of these models depends on the shape of the tensors, which makes it difficult to optimize the tensors before the model runs. There are two common solutions to this problem. The first is to add useless data to the input to match a pre-optimized tensor library. The second is to use small basic tensors to create a tensor that is closest in size to the input data and then tune it to minimize padding. However, this second solution can be time-consuming.

This paper proposes a new technique for deep learning compilers called FTuner. Instead of using a large design space or training a cost model, we use an abstract computational unit called the uKernel to patch together small, various-sized tensors to match the shape of the input tensor. We determine the shape of the uKernel using an analytic hardware information model. Experiments show that the FTuner can achieve comparable operators and end-to-end performance to vendor libraries and achieves 3\% speedup on existing auto-tuner with the model-training compiler while reducing tuning time by two orders of magnitude.

\end{abstract}

\begin{IEEEkeywords}
dynamic shape tensor, deep learning compilation, tensor program, auto-tuning
\end{IEEEkeywords}

\section{Introduction}
Large AI models \cite{dalle,chatgpt,seg-gpt}, have complex structures, numerous parameters, and remarkable intelligence advantages. Therefore, improving the efficiency of these models on hardware has become crucial. However, achieving the theoretical performance of hardware can only be possible when deep learning models are finely tuned for this hardware. It is common practice to develop libraries for deep learning operators on different hardware manually, such as cuBLAS \cite{cublas}, cuDNN \cite{cudnn}, CUTLASS \cite{cutlass}, and oneAPI \cite{oneapi}. However, developing and maintaining these libraries can be costly and time-consuming due to the frequent iteration of the model and hardware.

Deep learning compilers\cite{tvm,tc,halide,mlir}, can generate high-performance programs for deep learning models on hardware. Some compilers, such as Ansor\cite{ansor}, even incorporate auto-tuners to optimize tensor programs automatically for optimal performance. However, the input shape of large models cannot be known during the compilation phase. Therefore, if the shape of tensors is known only at runtime instead of compile time, it cannot guarantee effective execution for tensors in all possible shapes. For instance, in speech recognition models like WaveNet \cite{wavenet}, the input speech duration can vary, and in natural language processing like BERT\cite{bert}, the input sequence length ranges from a single word to hundreds. These scenarios involve what is known as dynamic shape tensors.

\begin{figure}
    \centering
    \includegraphics[width=\linewidth]{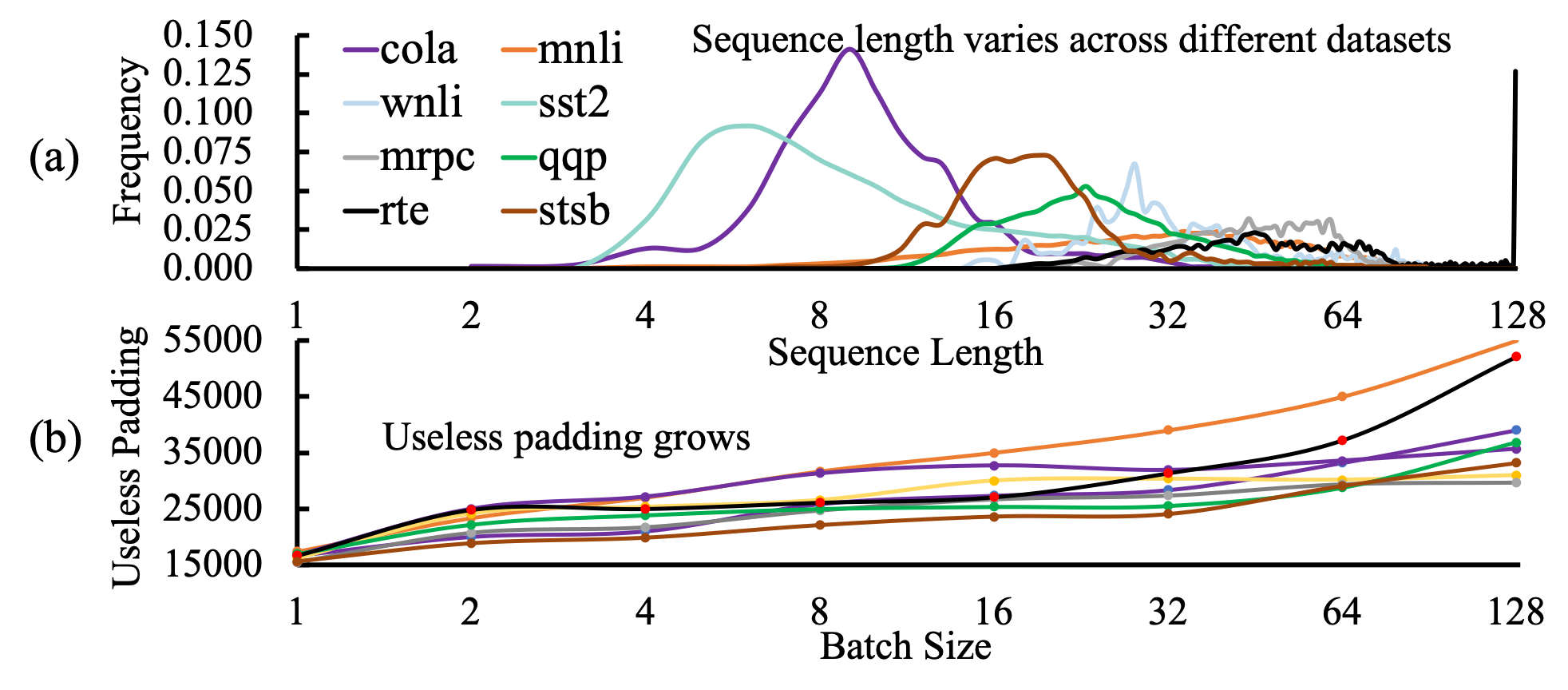}
    \caption {Tensor shape diversity and padding cost. (a) The shape of input tensors varies across different datasets from the standard NLP benchmark GLUE \cite{glue}. (b) As the batch size increases, the amount of useless padding of the batch matrix multiplication grows.}
    \label{fig:freq}
\end{figure}

\begin{figure}
    \centering
    \includegraphics[width=0.75\linewidth]{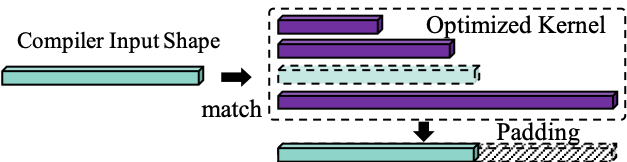}
    \caption{Illustration of Padding for Dynamic Shapes. Padding the dynamic shape to match the optimized kernel in the manual library.}
    \label{fig:pad}
\end{figure}

The problem of dynamic shape is quite common. In Fig. \ref{fig:freq} (a), we can see the variety of shapes in terms of sequence length in an NLP benchmark called GLUE\cite{glue}. Currently, the compiler can only choose the smallest basic-shape tensor from the available library and match it with additional padding data, as shown in Fig. \ref{fig:pad}. However, using padding data may lead to a significant reduction in performance. Large-scale AI models typically use a large batch size, and Fig. \ref{fig:freq} (b) illustrates how padding to the batch matrix multiplication (Matmul) will gradually expand as the batch size increases.

Recent studies on auto-tuning dynamic shape tensors, such as DietCode\cite{dietcode} and HAOTuner \cite{hao}, aim to reduce padding by creating small task units that can compose dynamic shape tensors. Compilers usually employ cost models to select high-performance units for dynamic shape tensors. However, training these models is time-consuming since these units are randomly generated and numerous in quantity. Roller \cite{roller}, on the other hand, does not require any cost model but predicts tensor performance through strict hardware alignment. However, this method is only effective when the shape is known before the running. When dealing with dynamic shapes, Roller still generates a significant amount of padding, making it challenging to outperform the cost model training compilers.

To address this issue, we propose FTuner, a tensor tuning technique that enables the fast generation of high-performance kernel code for dynamic tensors. We introduce a novel abstract computing unit called uKernel and constrain its generation through hardware features and multiple metrics. Based on the metrics retained in uKernel, synthesis index analysis can help rapidly find high-performance uKernel-based programs. During the runtime of tensor programs, FTuner combines different uKernels to reduce padding in dynamic shape tensor programs. We have evaluated FTuner on both standard deep learning benchmarks and emerging new workloads against vendor libraries and state-of-the-art compilers, using a wide range of input shapes. Experiment results show that FTuner performs comparably to manual libraries on nearly half of the shapes and can be portable to different architectures.

In summary, this paper makes the following contributions:
\begin{itemize}
    \item  We propose an abstract computing unit called uKernel to create dynamic tensor programs. The size of the uKernel can be adjusted according to different memory hierarchies and computing resources instead of a time-consuming cost model. 
    \item We implement a compiler framework FTuner based on uKernel. The compiler generates high-performance uKernels for dynamic shape tensors during the compilation phase. During runtime, FTuner combines different uKernels to create programs with small padding.
    \item The evaluation has shown that the techniques used by FTuner outperform state-of-the-art compilers on dynamic tensors. Furthermore, compared to compilers with model-training, FTuner reduces compilation time by two orders of magnitude, while achieves speedup by 3\% on typical operators.
\end{itemize}

This paper is organized as follows. \S \ref{section2} provides the background and motivation of this paper. \S \ref{section3} gives an overview of  FTuner. \S \ref{section4} introduces the compilation phase of FTuner. \S \ref{section5} introduces the runtime of FTuner. We evaluate FTuner in \S \ref{section6} and conclude this paper in \S \ref{section8}.

\section{Motivation}
\label{section2}

\begin{figure}
    \centering
    \includegraphics[width=\linewidth]{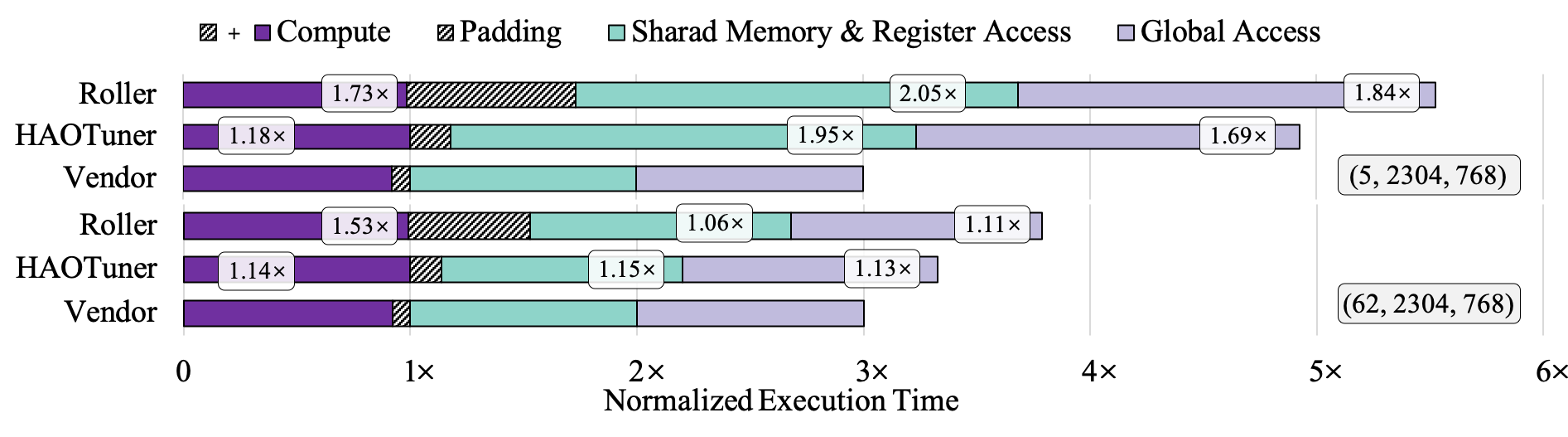}
    \caption{The execution time breakdown of tensors optimized by different compilers. We used the NCU \cite{ncu} to estimate the computation and memory access times for two shapes of the Dense operator on V100, normalized by Vendor. Roller \cite{roller} has a higher proportion of padding. \textbf{Since the padding time of the Vendor cannot be measured, we estimate the padding time roughly by calculating the difference in computation time between the current kernel and the strictly aligned kernel.}}
    \label{fig:insight-ncu}
\end{figure}

\subsection{The Gap Between Current Auto-tuner and Dynamic Shape Tensors Requirement}
The current auto-tuners for compilers, such as Ansor \cite{ansor}, theoretically can explore the entire solution space and find the best schedule for tensor programs across all possible shapes. However, it's important to consider the significant time and computational power costs associated with this approach. These compilers utilize a cost model that employs machine learning, such as XGBoost \cite{xgb}, Chameleon \cite{chameleon}, AdaTune\cite{adatune}, and heuristic search techniques like simulated annealing \cite{sa} and evolutionary algorithms\cite{ea} to find high-performance programs from a large solution space. Ansor, for instance, takes about 19.3 hours on V100 GPU to optimize only 8 shapes of input tensors for all the operators of the end-to-end model of BERT. Instead of creating optimal tensor programs with cost models, Roller \cite{roller} achieves highly predictable performance of tensor programs through strict hardware alignment. However, its method for handling dynamic shapes is not optimal.

DietCode\cite{dietcode} has an abstract computing unit that acts as a link between the tensor program and the hardware resource. By dividing the dynamic shape tensor input's kernel into multiple units and scheduling them onto the CUDA cores of the GPU, the program can be tuned by manipulating the units. To address performance instability issues across different devices in DietCode, HAOTuner \cite{hao} uses transfer algorithms to reduce the training time of the cost model slightly. However, it still takes 3 hours to optimize the entire BERT network and 35 minutes to optimize a single Dense operator. HAOTuner has a similar technique to DietCode and performs better performance than DietCode.

\begin{figure}
    \centering
    \includegraphics[width=0.95\linewidth]{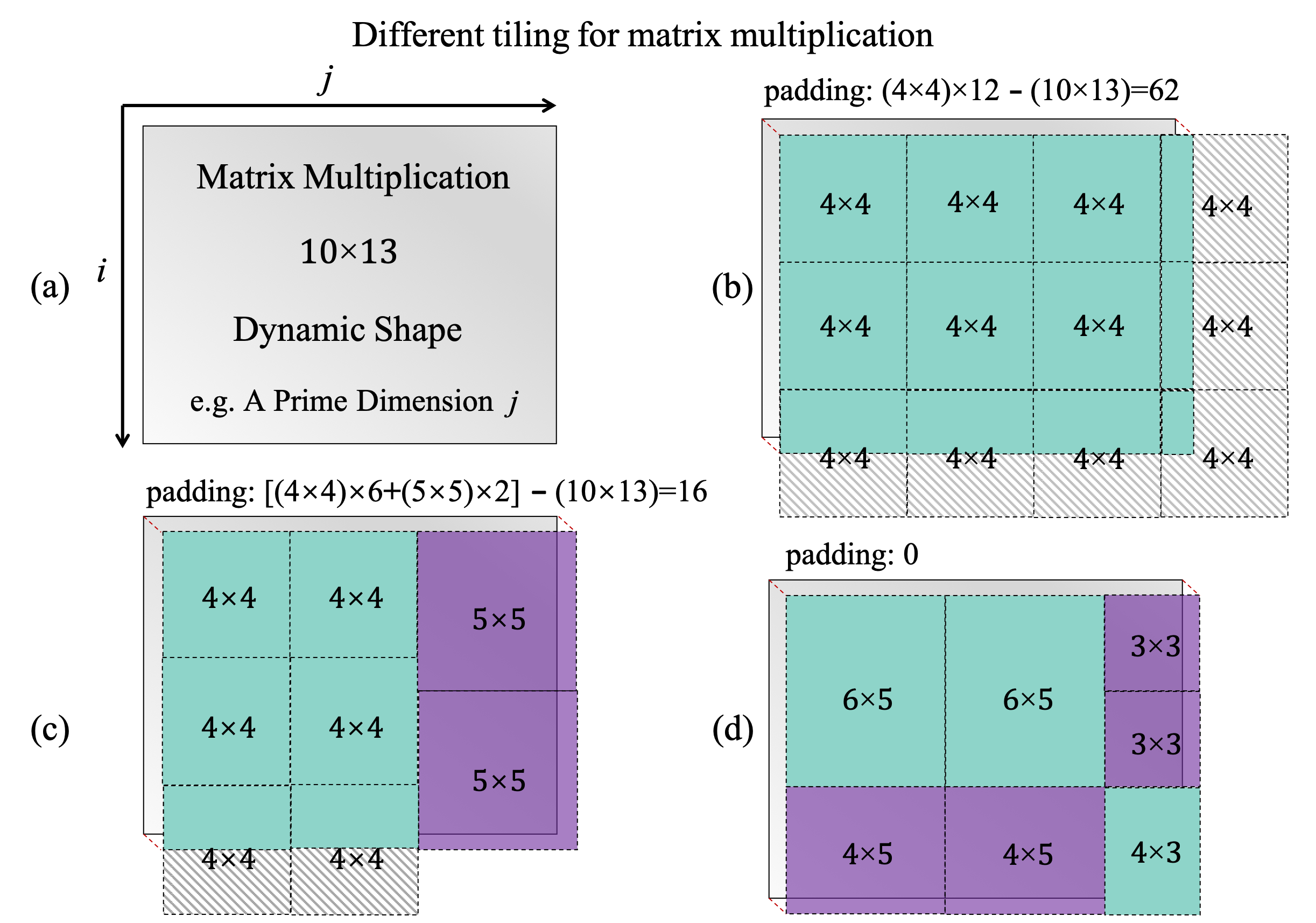}
    \caption{Different tiling for matrix multiplication. (a) represents a matrix multiplication output dimension with a prime size axis, (b) is composed of a single kernel, (c) is the method adopted in this paper, which achieves zero-padding along the j-axis using two kernels, and (d) represents an ideal combined state that is difficult to achieve.}
    \label{fig:shape}
\end{figure}

\subsection{Observation and Insights}
Although HAOTuner \cite{dietcode} can perform relatively well on dynamic shape tensors, it takes too long to complete the tuning process. Roller \cite{roller} can significantly reduce the tuning time, but its support for dynamic shapes is unsatisfactory. To investigate this issue, we analyzed the computation and memory access time of Roller. We found that the performance degradation is mainly due to padding, as shown in Fig. \ref{fig:insight-ncu}. Padding introduces additional boundary checks in nested loops, which slows down the computation process. Existing techniques have limitations in accurately profiling the computation time at a fine-grained level since computation and memory access time overlap. We tried multiple approaches, including tracking the number of instructions executed, but all exhibited significant errors. To estimate the timing of each component, we utilized the clock cycle statistics from NVIDIA NCU Tool \cite{ncu}. On average, the Roller exhibits 1.62$\times$ longer computation time than the Vendor. Therefore, if we can obtain abstract computing units like Roller, we can significantly reduce the tuning time. To enhance the performance of this approach, we need to minimize the program padding.

We found that using different basic tensor computation units to combine dynamic shape tensors can reduce padding. As shown in Fig. \ref{fig:shape}, (a) is a Matmul operator with a dynamic shape. HAOTuner uses only one unit to compose this tensor, as shown in (b). (c) can almost eliminate padding and enable using units with higher compute-to-memory ratios, but the challenge lies in accurately selecting these units. (d) represents an ideal state that we have not yet achieved. It motivates us to design a novel abstract computing unit and FTuner to select it and reduce padding.

\begin{figure}
    \centering
    \includegraphics[width=\linewidth]{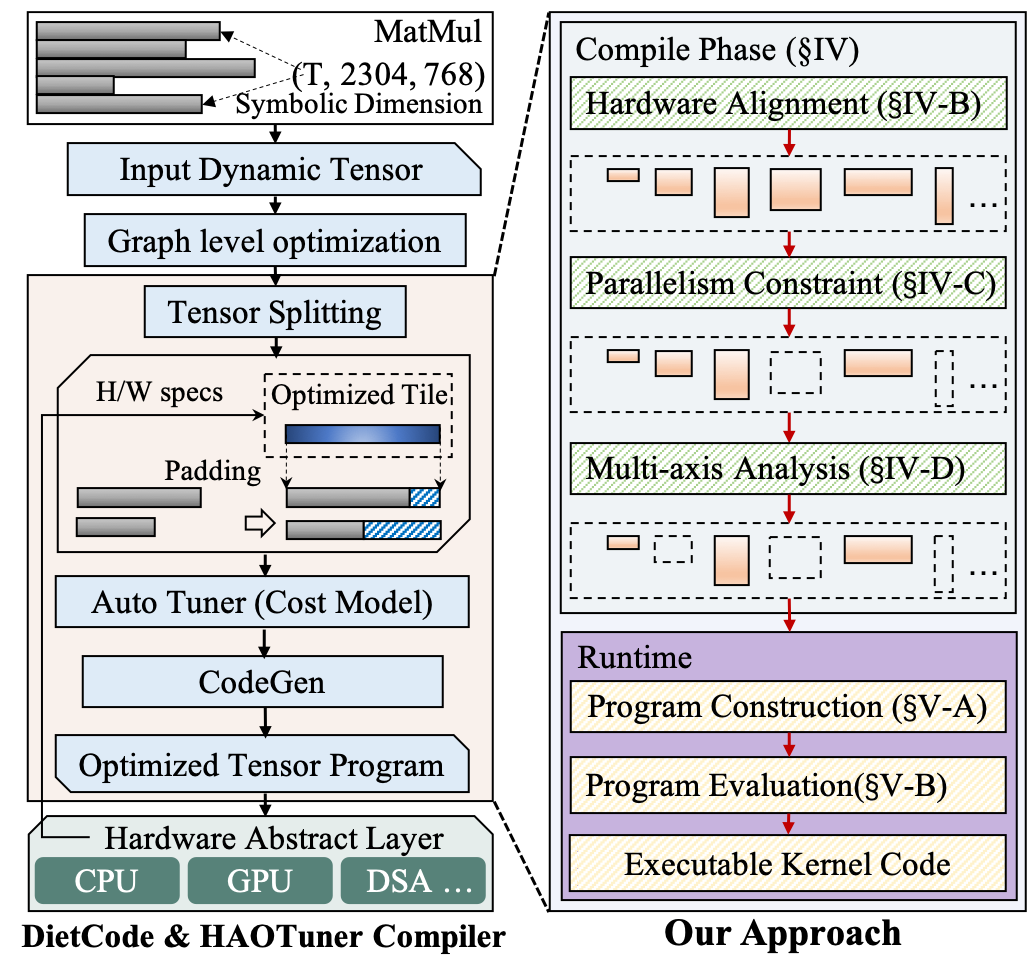}
    \caption{The overall architecture of FTuner. Taking the Matmul operator as an example, we assume that the input shape is denoted by the symbol T along the i-axis, while the j and k axes remain fixed. We replaced the portion from tensor splitting to generating optimized tensor programs in DietCode \cite{dietcode}.}
    \label{fig:pipeline}
\end{figure}

\section{FTuner Design Overview}
\label{section3}
The main idea of FTuner is to use a small unit tensor, which we name ukernel, as the building block to create target variable input tensors. An analytical model uses the unit tensor. The same input tensor can be composed of uKernels of different sizes. The distinction between uKernels lies in the number of iterations in each layer of loops, which corresponds to different numbers of thread blocks. To minimize the useless padding, FTuner matches input tensors with multiple-sized ukernels. The uKernel is a predefined operator template, which is implemented as a TVM \cite{tvm} schedule with its built-in scheduling primitives. FTuner can find a set of high-performance uKernels ahead of running time. We compare the overall architecture of FTuner with Dietcode \cite{dietcode} and HAOTuner \cite{hao} in Fig. \ref{fig:pipeline}. HAOTuner has a similar technique to DietCode and performs better performance than DietCode. We adopted the same tuning process as DietCode while replacing the portion from tensor splitting to generating optimized tensor programs.

\begin{figure}
    \centering
    \includegraphics[width=\linewidth]{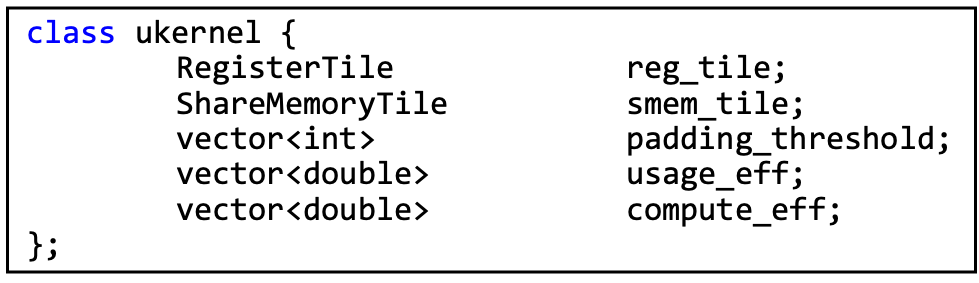}
    \caption{The data structure of uKernel. It is an abstract computing unit representing a fraction of the tensor program. To facilitate the evaluation of uKernel's performance, we define the tile size at two memory hierarchies and three performance metrics.}
    \label{fig:ukernel-code}
\end{figure}

Given an operator's tensor program (including computational rules and specific input shape), a uKernel represents a code segment constituting that tensor program. It determines the scope of tensor computations under the current computational rules. To enhance the performance of uKernels, we establish performance parameters for each one. Fig. \ref{fig:ukernel-code} shows the data structure of ukernels, including the tile sizes at different memory hierarchies, the proportion of effective computations, resource utilization, and the compute-to-memory ratio. Tab. \ref{tab:ukernel} lists the keywords used in the data structure. The uKernel resides in the streaming multiprocessors (SMs) of the GPUs. Each SM consists of several thread blocks of different sizes, and uKernel is abstracted as a thread block.

\begin{table}[H]
    \centering
    \renewcommand\arraystretch{1.1}
    \caption{Keyword in uKernel.}
    \resizebox{\linewidth}{!}{
        \begin{tabular}{c|c}
        \hline
        \textbf{Keyword} &\textbf{Definition}\\
        \hline
        reg\_tile &The tile size of uKernel in the register level.\\
        \hline
        smem\_tile &The tile size of uKernel in the shared memory level.\\
        \hline
        padding\_threshold &The proportion of the part without padding in uKernel.\\
        \hline
        usage\_eff &The resource utilization of uKernel.\\
        \hline
        compute\_eff &The compute-to-memory ratio of uKernels.\\
        \hline
    \end{tabular}
    }
    \label{tab:ukernel}
\end{table}

We implemented FTuner based on TVM \cite{tvm}. The front-end input of FTuner is an ONNX graph \cite{onnx}, which utilizes subgraph fusion from Ansor \cite{ansor} and operator fusion from Relay \cite{relay}. The fused operators could be presented as multiple nested iterations, and FTuner can optimize each simultaneously. FTuner exports tensor expressions \cite{halide} and generates a uKernel set through its filter algorithm for the operator extracted from the optimized graph.

\begin{figure*}
    \centering
    \includegraphics[width=\linewidth]{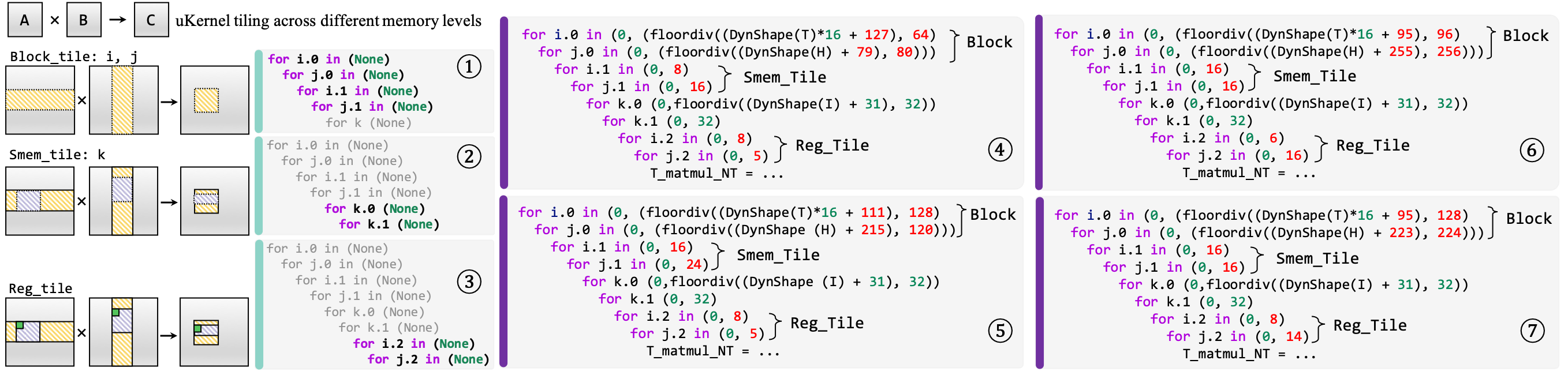}
    \caption{Representing the actual tiling of uKernel at different levels using code. \ding{172}$-$\ding{174} shows uKernel tiling across the different levels. This example applies to operators represented by nested loops. The axes space(i,j) and reduce(k) can also represent the loop of the convolution. Such as space(n,k,p,q) and reduce(c,r,s). \ding{175}$-$\ding{178} provides four different loop states of uKernel. i.2 and j.2 represent the register tiles, i.1 and j.1 indicate the shared memory tiles, and also represent thread tiles. i.0 and j.0 represent block-level tiles.}
    \label{fig:tile-code}
\end{figure*}

During the compilation stage, FTuner initiates a group of ukernel candidates for all the possible shapes of input tensors and creates an operator template where the type of ukernels is parameterized. At the runtime stage, FTuner constructs multiple candidates of the tensor programs for the input tensors and chooses the optimal one to execute. We introduce the compilation phase and running stage in Section 4 and Section 5, respectively.

\section{Compilation stage of FTuner}
\label{section4}
During the compilation phase, FTuner aims to generate a high-performance uKernel set. We use input dynamic shape tensor as workload, which includes every possible shape. We introduce the uKernel definition and then initialize the uKernel set through three parts: hardware alignment, parallelism constraints, and multi-axis analysis.

\begin{equation}
    \begin{split}
        SIA\;Score = c\_{0}CMR + c\_{1}Pad + c\_{2}Occ
        \label{eq:sia}
    \end{split}
\end{equation}

We systematically derive each term of Equation \ref{eq:sia} for uKernels step by step. Equation \ref{eq:sia} serves as an analytical model used at the runtime stage to evaluate the performance of programs. Here, CMR, Pad, and Occ denote compute\_eff, padding\_threshold, and usage\_eff within the uKernels, respectively. These three factors are calculated using Equations \ref{eq:mem}, \ref{eq:pad}, and \ref{eq:occ}, respectively.

\subsection{Hardware Alignment of uKernel}
\label{section4.2}
To align with hardware, FTuner tiles the computation from two memory hierarchies. For a more intuitive illustration, as shown in the left of Fig. \ref{fig:tile-code}, the naive loop for Matmul $C\_{i,j}=\sum\_{k}A\_{i,k}B\_{k,j}$ as an example. In thread block, the loop structure in \ding{172} represents the tiling of shared memory along the i and j axes. \ding{173} represents the tiling along the k-axis. The tiling of the shared memory is completed only when the k-axis is finished. \ding{174} illustrates the register tiles, where each tile along the k-axis comprises multiple registers.

FTuner provides the RegTileRule() (alg. \ref{alg1} line \ref{alg1:line:2}) to tile reg\_tile based on factors of the axis length. Calculate all factors of the dynamic axis input. These factors serve as an array of the tiling for the register. For example, when the sequence length is a prime number (53), we store the factors of adjacent numbers (52, 54) in the array. SmemTileRule() (alg. \ref{alg1} line \ref{alg1:line:3}) for shared memory tiling. First, determine the major axis of the loop, typically row-major or col-major. smem\_tile is first tiled along with the major axis, and the remaining axes grow linearly based on reg\_tile. For example, the k-axis is aligned with a multiple of 8, whereas for the i and j axes, we compute the factors that can evenly divide them. For float32, we select multiples of 8 on NVIDIA GPUs. For different datatypes, depending on the width of memory transactions, different alignment multiples can be used. Finally, the scale-up is repeated to expand the smem\_tile. This rule obtains an increasing smem\_tile while traversing all reg\_tiles.

Comparing \ding{175} and \ding{176}, the same register tile results in different shared memory tiles (also represent thread-level tiles). Since the thread allocation determines the size of shared memory requested by the uKernel, the thread-level tile indicates the alignment of the uKernel with shared memory. \ding{177} and \ding{178} show that different register tiles can also result in the same shared memory tile. We aim to explore all possible tiling options to avoid missing high-performance uKernels. This phase on V100 can generate approximately 2 million uKernels for 128 input shapes.

\begin{table}[H]
    \centering
    \renewcommand\arraystretch{1.1}
    \caption{Important Notations in \S \ref{section4.3}.}
    \resizebox{\linewidth}{!}{
        \begin{tabular}{c|c}
        \hline
        \textbf{Notations} &\textbf{Definition}\\
        \hline
        wkl &The workload represents the operator with shape.\\
        \hline
        K.pad, $\epsilon$ &The value of padding\_threshold K.pad is $\epsilon$.\\
        \hline
        wkl.base(K) &The bytes to compose the workload using K.\\
        \hline
        wkl.pad(K) &The bytes need to be padded.\\
        \hline
        K.occ, $\lambda$ &The value of usage\_eff K.occ is $\lambda$.\\
        \hline
        wkl/K &The number of uKernels required in the workload.\\
        \hline
        NumCores &The total number of SMs.\\
        \hline
        block.bound &Thread blocks allocated to one SMs.\\
        \hline
        $\zeta$ &The default active thread block per SM.\\
        \hline
        K.wkl.block &Active thread blocks per SM required by the workload.\\
        \hline
        K.RegsInBlock &The registers in the thread block.\\
        \hline
        RestRegs &The registers required for the remaining variables.\\
        \hline
        REGS\_PER\_SM &The available number of registers per SM.\\
        \hline
    \end{tabular}
    }
    \label{tab:filter}
\end{table}

\subsection{Parallelism Constraint of uKernel}
\label{section4.3}
\textbf{Trade-off with padding \& occupancy.} Resource occupancy is a crucial index for parallelism, but the key focus of this subsection is how to select a uKernel with less padding but higher resource occupancy. Our filter algorithm (alg. \ref{alg1}) explores the padding and resource occupancy for each uKernel to provide an equal opportunity. Note that, three functions in alg. \ref{alg1} are executed sequentially, each returning a uKernel set defined as K.Align, K.Cross, and K.Filter, respectively. We employ the CrossPick algorithm (line \ref{alg1:line:7}) to select uKernels (K) with higher parallelism. Tab. \ref{tab:filter} lists the notations used in \S \ref{section4.3}.

\begin{small}
    \begin{equation}
        \begin{split}
            K.pad = \frac{wkl.base(K)}{wkl.pad(K)+wkl.base(K)}
            \label{eq:pad}
        \end{split}
    \end{equation}
\end{small}

\begin{small}
    \begin{equation}
        \begin{split}
            K.occ = \frac{wkl/K}{ceil\_by(wkl/K, \; NumCores)}
            \label{eq:occ}
        \end{split}
    \end{equation}
\end{small}    

\begin{small}
    \begin{equation}
        \begin{split}
            K.RegsInBlock \leq & \frac{REGS\_PER\_SM}{block.bound}\\
            where \; K.RegsInBlock=&(reg\_tile_{i,j}+RestRegs)\\
            &\times smem\_tile_{i,j}
            \label{eq:block}
        \end{split}
    \end{equation}
\end{small}

The padding\_threshold is calculated by \eqref{eq:pad}. A larger value of K.pad indicates less padding. The usage\_eff is calculated by \eqref{eq:occ}. As each uKernel is assigned to one SM, wkl/K is also related to the number of occupied SMs. K.occ denotes the hardware occupancy ratio of a wkl composed using K.


We set the K.pad to the minimum $\epsilon_{min}$ and the K.occ to the maximum $\lambda_{max}$. We gradually increase $\epsilon$ and decrease $\lambda$ by a certain proportion. Although we utilized the same stride size across different GPUs, it is a customizable value. We observed that the $\epsilon$ (K.pad) and $\lambda$ (K.occ) of uKernels are generally inversely proportional. Furthermore, there exists a certain proportionality between these factors, which is hardware-dependent. For V100, we set the stride $\Delta\epsilon : \Delta\lambda$ to 10:1. For instance, $\epsilon_{min}=50\%$ increases by 1\% and $\lambda_{max}=95\%$ decreases by 0.1\% at each iteration. The filtering process terminates when either $\epsilon_{max}=95\%$ or $\lambda_{min}=90\%$. The "Next()" signifies moving to the next uKernel, while the function "Forward()" indicates the next stride. $\epsilon_{min, max}$, $\lambda_{min, max}$ and stride size are hyperparameters.

\begin{algorithm}[t]
    \caption{Filter Algorithm.}
    \label{alg1}
    \LinesNumbered
    \small
    \KwIn{Dynamic Shape Workload}
    \KwOut{uKernel Candidates}
    
    \SetKwFunction{FInit}{HardwareAlign}
    \SetKwFunction{FCrossPick}{CrossPick}
    \SetKwFunction{FSetBound}{SetBound}
	
    \SetKwProg{Fn}{Func}{:}{}
    \Fn{\FInit{shape:wkl.DynShape, Dev:device}}{ \label{alg1:line:1}
        reg\_tile = RegTileKernel(Shape)\; \label{alg1:line:2}
        smem\_tile = SmemTileKernel(Shape)\;  \label{alg1:line:3}
        padding\_threshold $\in$ [$\epsilon_{min}, \epsilon_{max}$]\;
        usage\_eff $\in$ [$\lambda_{min}, \lambda_{max}$]\;
        Return K.Align\;
    }

    \SetKwProg{Fn}{Func}{:}{}
    \Fn{\FCrossPick{K:uKernel, mem:K.MemCapacity, wkl:Workload}}{   \label{alg1:line:7}

        \ForEach{K in K.Align}{
            \If{mem.IsOverflow()}{
                Next(K) \;
            }
            \ForEach{$\epsilon_{i}, \lambda_{j}$}{
                \If{$\epsilon_{i} > \epsilon_{max}$ or $\lambda_{j} < \lambda_{min}$}{
                    Next(K) \;
                }
                \If{$K.pad \geq \epsilon_{i}$ and $K.occ \geq \lambda_{j}$}{
                    wkl.Add(K) \;
                }
                Forward($\epsilon_{i}, \lambda_{j}$) \;
            }
            Next(K) \;
        }
	Return K.Cross \;
    }

    \SetKwProg{Fn}{Func}{:}{}
    \Fn{\FSetBound{K:uKernel, wkl:Workload, Dev:device}}{  \label{alg1:line:19}
        block.bound = min(K.wkl.block, Dev.$\zeta$) \; \label{alg1:line:20}
        \ForEach{K in K.Cross}{
            \If{K.IsInBound()}{     \label{alg1:line:22}
                wkl.Add(K) \;
            }
        }
        Return K.Filter \;
    }
    
\end{algorithm}

\textbf{Parallelism bounds.} Increasing the use of registers in uKernel can improve hardware utilization, but register spillage can lead to severe performance degradation. FTuner sets parallelism bounds to control the number of thread blocks in one SM, as shown in SetBound (line \ref{alg1:line:19}). We retain default parameters $\zeta$ obtained from heuristic search, whose value is 2 on V100. We set the block.bound to the minimum of $Dev.\zeta$ and K.wkl.block (line \ref{alg1:line:20}). This is because, for large workloads, choosing $Dev.\zeta$ can avoid activating too many blocks, which may lead to register spillage. For small workloads, select K.wkl.block allows for an even distribution of blocks across each SM. $Dev.\zeta$ denotes the default active thread block per SM in Device. K.wkl.block denotes the active blocks per SM required by the workload using uKernel K.

Next, we examine whether uKernels satisfy the constraints of parallel boundaries. The inequality for registers usage boundary judgment in K.IsInBound() (line \ref{alg1:line:22}) is shown in \eqref{eq:block}. RestRegs represent the registers required for the remaining variables generated during thread computation.

\begin{table}[H]
    \centering
    \renewcommand\arraystretch{1.1}
    \caption{Important Notations in \S \ref{section4.4}.}
    \resizebox{\linewidth}{!}{
        \begin{tabular}{c|c}
        \hline
        \textbf{Notations} &\textbf{Definition}\\
        \hline
        $\Phi_{wkl}$ &The size of workload on space axis.\\
        \hline
        $\phi_{K}$ &The tile size of uKernel on the shared memory.\\
        \hline
        Active Blocks &The total blocks activated across all computing units.\\
        \hline
        $\psi$ &The threshold of compute-to-memory ratio.\\
        \hline
        wkl.FLOPs &The floating-point operations of workload.\\
        \hline
        FLOPS &The floating-point operations speed of the hardware.\\
        \hline
        $K_{mem}$ &Memory access latency of uKernel.\\
        \hline
        $data_{R}$ &Amount of data read from global memory.\\
        \hline
        $data_{W}$ &Amount of data written to global memory.\\
        \hline
        $data_{transW}$ &Amount of data written to shared memory.\\
        \hline
        $data_{transR}$ &Amount of data read from shared memory.\\
        \hline
        $bw_{G}$ &Bandwidth of global memory.\\
        \hline
        $bw_{S}$ &Bandwidth of shared memory.\\
        \hline
        \end{tabular}
    }
    \label{tab:axis}
\end{table}

\subsection{Multi-axis Analysis of uKernel}
\label{section4.4}
After the parallelism constraint of uKernel, FTuner needs to confirm two things: 1) whether using this uKernel can saturate the computing resources for the workload, and 2) whether the uKernel is computationally intensive. FTuner selects uKernels that simultaneously satisfy these two conditions. Then, we assess two conditions from space and reduce axes.


\begin{figure*}
    \centering
    \includegraphics[width=\linewidth]{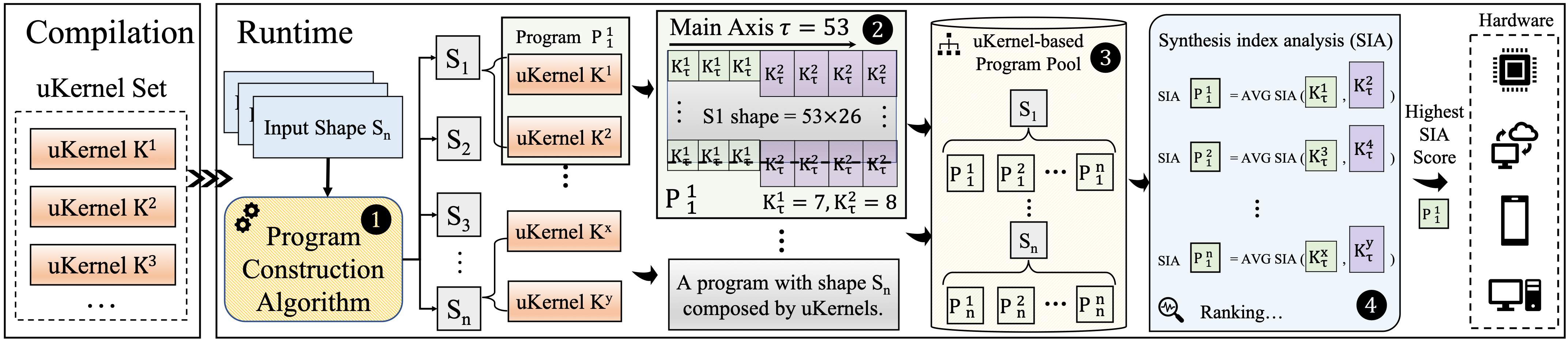}
    \caption{The FTuner at runtime. Combination algorithm \ding{182} generates multiple uKernel-based programs $P_{n}$ for each input shape $S_{n}$. In \ding{183}, $P_{1}^{1}$ is composed of uKernel $K^{1}_{\tau}$ and $K^{2}_{\tau}$, which can perfectly divide the main axis of shape $S_{1}$. Each shape has many uKernel-based programs in the program pool \ding{184}. We rank these programs by SIA score in \ding{185}.}
    \label{fig:infer}
\end{figure*}

\textbf{Space axis analysis.} The utilization of computing resources by the workload is determined based on the number of computing units and the number of thread blocks activated in each computing unit. FTuner set \eqref{eq:space} to determine whether the uKernel satisfies the saturation condition.

\begin{small}
    \begin{equation}
        \begin{split}
            &\frac{\Phi_{wkl}}{\phi_{K}} \geq Active \; Blocks\\
            \label{eq:space}
        \end{split}
    \end{equation}
\end{small}

\begin{small}
    \begin{equation}
        \begin{split}
            \left[ \frac{wkl.FLOPs}{FLOPS} :
            K_{mem}\right] \geq \psi\\
            \label{eq:reduce}
        \end{split}
    \end{equation}
\end{small}

\begin{small}
    \begin{equation}
        K_{mem}=max(\frac{data_{R}}{bw_{G}}+\frac{data_{W}}{bw_{G}},\frac{data_{transW}}{bw_{S}}+\frac{data_{transR}}{bw_{S}})\\
        \label{eq:mem}
    \end{equation}
\end{small}

$\frac{\Phi_{wkl}}{\phi_{K}}$ represents the number of uKernels composing the workload as shown in Tab. \ref{tab:axis}, and if it satisfies \eqref{eq:space}, it indicates computational saturation. The Active Blocks are equal to the product of ACTIVE\_BLOCK\_PER\_SM and NumCores. Note that, the V100 has 80 SMs, and when the number of blocks is 81, it leads to poor performance. \S \ref{section4.3} has excluded the "slightly larger than Active Blocks" uKernel through K.occ \eqref{eq:occ}, as we set ACTIVE\_BLOCK\_PER\_SM as 2, indicating that each SM is active with 2 blocks simultaneously. This setting is close to the maximum parallelism, so we don’t consider the case with fewer blocks.

\textbf{Reduce axis analysis.} FTuner determines whether uKernel is compute-intensive using \eqref{eq:reduce}. The part before the inequality represents the compute-to-memory ratio of the uKernel, denoted by compute\_eff. If compute\_eff surpasses the threshold $\psi$, it suggests that the uKernel is computation-intensive. The memory access latency of uKernel K.mem is computed by \eqref{eq:mem}.

\section{Runtime stage of FTuner}
\label{section5}
After section IV, we obtained a high-performance uKernels set. Due to the differences in hardware and input shape, the number of uKernels varies from twenty to hundreds.

\subsection{Program Construction}
\label{section5.1}
In this subsection, FTuner aims to find a combination of uKernels to generate programs with minimal padding. For instance, when one axis of shape is a prime number, it is challenging to achieve divisibility with just one size of uKernel. Non-divisibility leads to many boundary checks during memory access, reducing program performance.

To provide more combination options, we select the axis with max size as the main axis $\tau$ as shown in Tab. \ref{tab:runtime}. Alg. \ref{alg2} aims to find a uKernel combination that perfectly covers the $\tau$-axis. For a problem size H(), there exist two uKernels, denoted as $K^{1}$ and $K^{2}$, that can perfectly divide H, resulting in $H=(Num_{1} \times K_{\tau}^{1}, Num_{2} \times K_{\tau}^{2})$. For example,as shown in Fig. \ref{fig:infer} \ding{183}, if H=53, $K_{\tau}^{1}=7$, $K_{\tau}^{2}=8$, and $H=(3 \times K_{\tau}^{1}, 4 \times K_{\tau}^{2})$. $P_{1}^{1}$ is one of the uKernel-based programs found by algorithm \ding{182} for input shape $S_{1}$. As shown in Alg. \ref{alg2:line:14}, we do not exclude using tensor programs generated by a single uKernel. We retain uKernels that are divisible by the main axis size H and store their programs in a pool (uProg can see in Fig. \ref{fig:infer} \ding{184}).

In the CombinSearch(), we iterate through all uKernels. First, we check if $K_{\tau}$ is a factor of H (line \ref{alg2:line:5}). For example, since $K_{\tau}=7$ is not a factor of H=53, we continue searching for the next uKernel that can form 53. In this case, H is reduced by $K_{\tau}$ in the next search round, making H=46 as shown in line \ref{alg2:line:10}. This process continues until we find the perfect division, i.e., H=(7,8), stored in the CombinSet, as shown in line \ref{alg2:line:18}. The combination algorithm ensures that programs generated by FTuner require no padding in at least one axis.

\begin{algorithm}[t]
    \caption{Combination Algorithm.}
    \label{alg2}
    \LinesNumbered
    \small
    \KwIn{uKernel Set after \S \ref{section4.4}}
    \KwOut{uKernel-based Tensor Program}
    
    \SetKwFunction{FCombinSearch}{CombinSearch}

    \SetKwProg{Fn}{Func}{:}{}
    \Fn{\FCombinSearch{$K_{\tau}:K.dim.\tau, H:wkl.dim.\tau$}}{ \label{alg2:line:1}
        \ForEach{K in uKernel set}{
            \If{$H \leq 0$}{Next(K) \;}
            \eIf{$K_{\tau} \in divisor(H)$}{ \label{alg2:line:5}
                CombinSet.Add($K_{\tau}$) \;
                Next(K) \;
                Add uKernel to the combination set.
            }{
                
                CombinSearch($K_{\tau}, H -= K_{\tau}$) \; \label{alg2:line:10}
            }
        }
        Return  CombinSet\;
    }

    \ForEach{K in K.Filter}{
        \eIf{$K_{\tau} \in divisor(H)$}{
            uProg.Add($K_{\tau}$) \;  \label{alg2:line:14}
            The program is generated by only one uKernel.
        }{
            CombinSearch() \;
            uProg.Add(CombinSet) \; \label{alg2:line:18}
        }
        Return uProg \;
    }
\end{algorithm}

\begin{figure*}
    \centering
    \includegraphics[width=\linewidth]{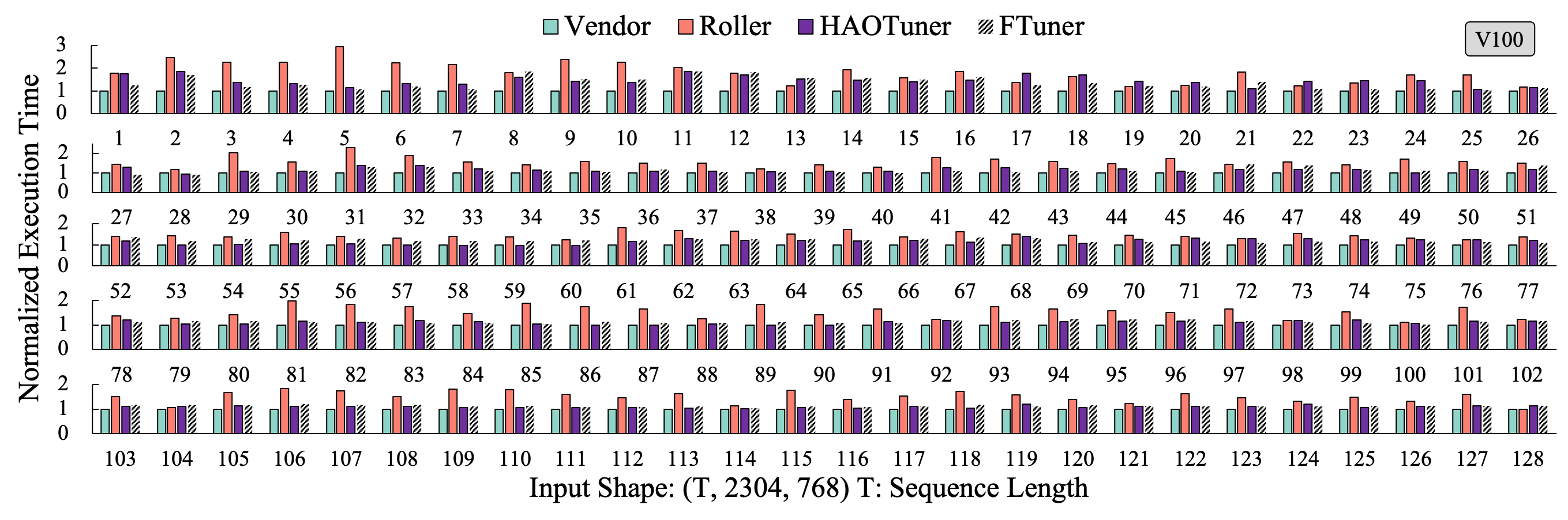}
    \caption{The execution time of the Dense operator on the V100. The computed definition of Dense is $C_{i,j}=\sum_{k}A_{i,k}B_{k,j}$. The Vendor normalized the values of 128 shapes. FTuner can rival Vendor on 44\% of the shapes. FTuner is up to 68\% better than Roller when T=5, 23\% on average. FTuner outperforms the HAOTuner in 59\% of the shapes. FTuner is up to 30\% better than HAOTuner when T=27, 0.6\% on average.}
    \label{fig:denseV100}
\end{figure*}

\begin{table}[H]
    \centering
    \renewcommand\arraystretch{1.1}
    \caption{Important Notations in \S \ref{section5}.}
    \resizebox{\linewidth}{!}{
        \begin{tabular}{c|c}
        \hline
        \textbf{Notations} &\textbf{Definition}\\
        \hline
        $\tau$, H &The maximum axis $\tau$ with size H.\\
        \hline
        $K_{\tau}^{1}$, $K_{\tau}^{2}$ &The size of uKernel $K^{1}$ and $K^{2}$ along the main axis $\tau$.\\
        \hline
        divisor(H) &The factors set of H.\\
        \hline
        $P_{n}^{1}$ &The 1st uKernel-based program for input shape $S_{n}$.\\
        \hline
        CMR, Pad, Occ &comput\_eff, padding\_threshold, usage\_eff in ukernel.\\
        \hline
        $c_{s}, s=0,1,2$ &The penalty coefficient for the three metrics.\\
        \hline
    \end{tabular}
    }
    \label{tab:runtime}
\end{table}

\subsection{Program Evaluation}
\label{section5.2}
FTuner will evaluate programs in the uProg using a synthesis index analysis (SIA) to select the best-performing uKernel-based program for workloads of different input shapes. We calculate the SIA score using a weighted sum of three metrics (as shown in \eqref{eq:sia}): compute-to-memory ratio, padding, and resource utilization. According to the SIA score rank, all uKernel-based programs in the uProg are When calculating the SIA score for a program composed of two uKernels, and we take their average (as shown in Fig. \ref{fig:infer} \ding{185}).

We set coefficients c0=c1=c2=1. Users have the flexibility to customize these coefficients. Since all the above-mentioned information is cached in the data structure of each uKernel, real-time queries can be performed when calculating the SIA scores. Then, FTuner selects the Top10 tensor programs based on their SIA scores for different input shapes of the workload and compares them with the measurement-on-device results of all programs. \S \ref{section6.3} validates the accuracy of the SIA. Note that FTuner does not require measurement-on-device at runtime. Additionally, during the runtime, the program construction and evaluation time is significantly smaller compared to the inference time (by about three orders of magnitude). Therefore, we include the runtime overhead in the inference time.

Users can modify each coefficient based on the type of operator and model, hardware features, and input tensor shape. For example, users can use NVIDIA's nvprof to determine operators' computation and memory access patterns. If the operator is compute-intensive, the coefficient of CMR can be increased. If the computing resources (obtained by nvidia-smi) of the hardware device are limited, such as embedded devices, the coefficient of Occ can be increased. In summary, even if users do not have guidance, selecting default coefficients can still result in high-performance tensor programs.




\begin{figure*}
    \centering
    \includegraphics[width=\linewidth]{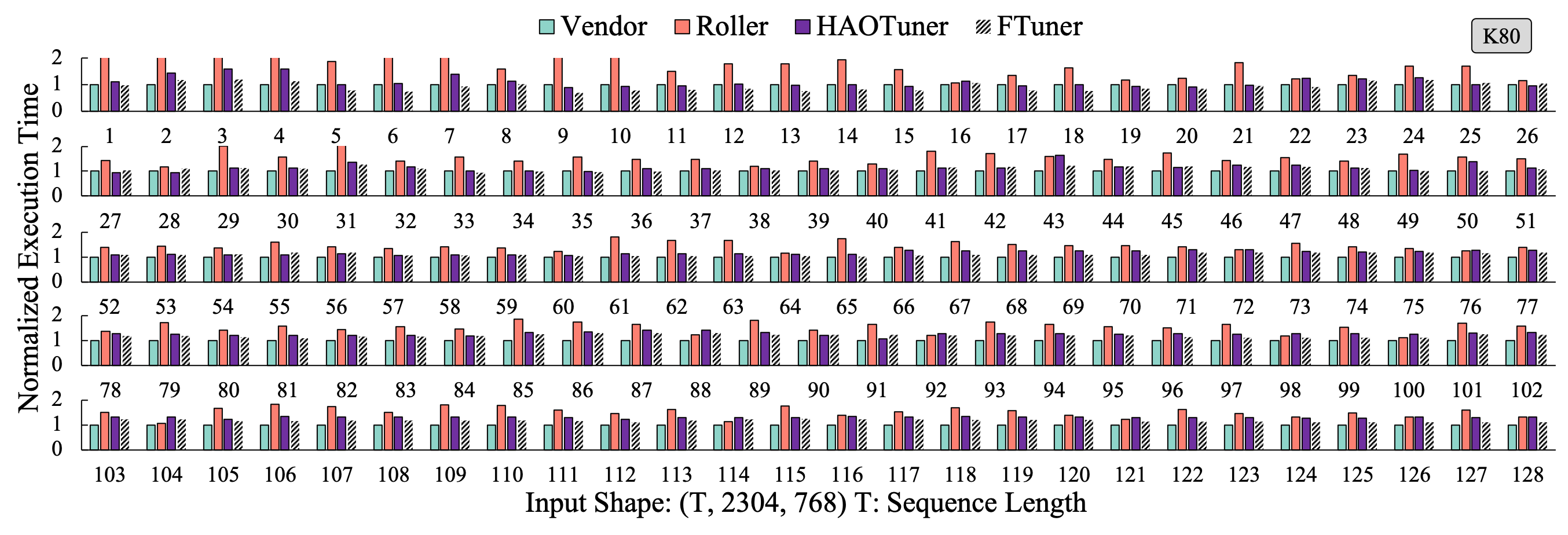}
    \caption{The execution time of the Dense operator on the K80. The computed definition of Dense is $C_{i,j}=\sum_{k}A_{i,k}B_{k,j}$. The Vendor normalized the values of 128 shapes. FTuner can rival Vendor on 50\% of the shapes. FTuner is up to 70\% better than Roller when T=9, 28\% on average. FTuner outperforms the HAOTuner in 98\% of the shapes. FTuner is up to 33\% better than HAOTuner when T=7, 8.2\% on average.}
    \label{fig:dense80}
\end{figure*}

\begin{figure*}
    \centering
    \includegraphics[width=\linewidth]{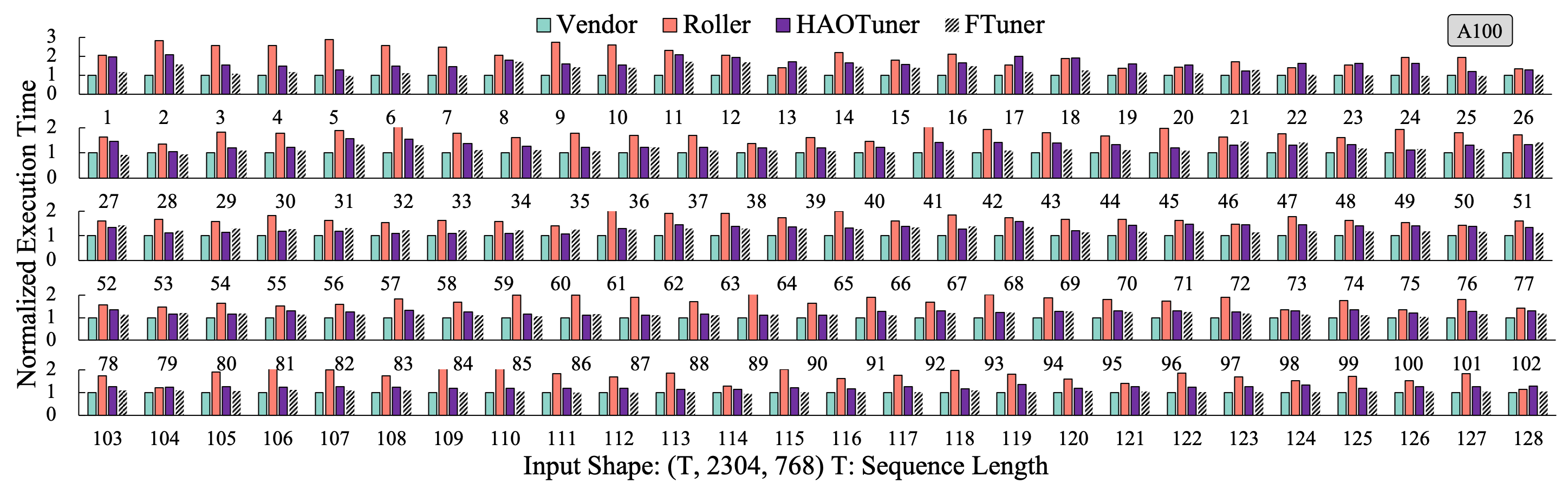}
    \caption{The execution time of the Dense operator on the K80. The computed definition of Dense is $C_{i,j}=\sum_{k}A_{i,k}B_{k,j}$. The Vendor normalized the values of 128 shapes. FTuner can rival Vendor on 40\% of the shapes. FTuner is up to 66\% better than Roller when T=5, 33\% on average. FTuner outperforms the HAOTuner in 88\% of the shapes. FTuner is up to 42\% better than HAOTuner when T=17, 12\% on average.}
    \label{fig:denseA100}
\end{figure*}

\section{Evaluation}
\label{section6}
We compare FTuner with three baselines:
1) Vendor, an operator library (cuBLAS \cite{cublas} for GPU). cuDNN \cite{cudnn} mainly optimizes convolution operators, and certainly, CUTLASS \cite{cutlass} demonstrates superior performance. Nonetheless, given the extensive number of experiments, we compared it with cuBLAS for convenience rather than manually implementing CUTLASS for each shape. To maintain consistency with the baseline, we did not use Tensor Cores. Tensorization is another work that is orthogonal to FTuner.
2) Roller \cite{roller}, the most representative fast compiler.
3) HAOTuner \cite{hao}, the state-of-the-art compiler for dynamic shape, with all evaluated search rounds set to 1000 trials. HAOTuner has a similar technique to DietCode and performs better performance than DietCode.

\textbf{Experimental setup.} We evaluated FTuner on five NVIDIA GPUs with different architectures: NVIDIA Tesla K80 (Kepler) and V100 (Volta), GeForce GTX 3090 and A100 (Ampere), and 4090 (Ada Lovelace). All of them used a 32-core Intel Xeon E5-2620 v4 @2.10GHz CPU. We mainly chose K80 and V100 to compare them with a very related work Roller \cite{roller}. We used CUDA 11.1 for K80; other devices used CUDA 11.4 and TVM (v0.8) as the software environment configuration.


\textbf{Workload.} We evaluated two models, BERT \cite{bert} and GPT-2 \cite{gpt2}, and two operators of attention \cite{Transformer}. FTuner tune operators that are performance bottlenecks, Dense and BatchMatmul, account for over 90\% of the invocations in these two models. For simple element-wise operators like ReLU and element-wise add, FTuner directly handles them using inline functions in PyTorch \cite{pytorch}. The computed definitions of Dense and BatchMatmul are from ONNX \cite{onnx}. We take the input length range of 1-128 for BERT as the experimental shapes for both single operators and end-to-end models (including GPT-2). We evaluate FTuner from four aspects: 1) Inference time for a single operator and end-to-end models, using Execution Time as the metric. 2) Compilation time. 3) Synthesis index analysis accuracy. 4) Padding and memory access analysis. To comprehensively evaluate the results of FTuner and to maintain consistency with HAOTuner, we evaluate BatchMatmul performance and the total optimization time of the compiler for 8 different shapes.

\textbf{Evaluation index.} We present the evaluation results in four ways:
1) For inference time for both operator and end-to-end models, We averaged the values from over one hundred runs in milliseconds (msec).
2) For compilation time, we report the time in seconds (sec).
3) For synthesis index analysis accuracy, we report the performance in GFLOPS.
4) For the padding and memory access analysis, we averaged the padding time in computation from over one hundred runs in milliseconds (msec).
The final results are calculated as the average across all devices. Except for the GFLOPS metric, lower is better.

We also experimented with models that include convolution operators. However, due to their relatively fixed shape, these operators were not included in the comparison between HAOTuner \cite{hao} and DietCode \cite{dietcode}. Therefore, we did not include results for convolutions in this paper.

\begin{figure}
    \centering
    \includegraphics[width=\linewidth]{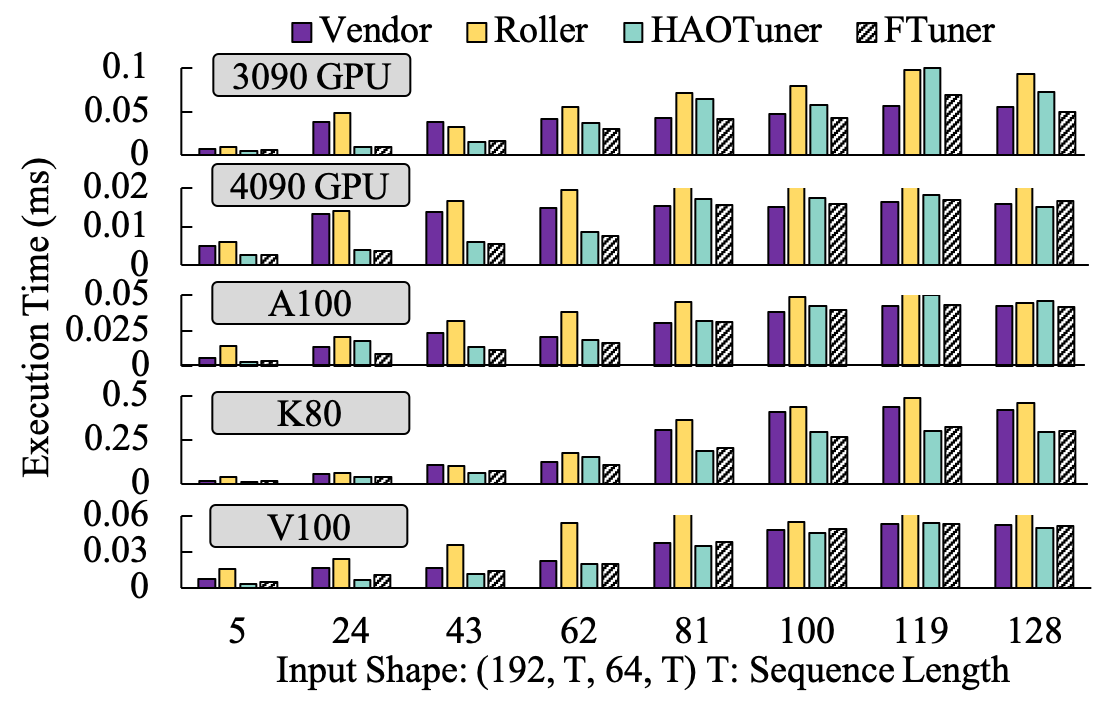}
    \caption{The execution time of the BatchMatmul. The computed definition of BatchMatmul is $C_{b,i,j}=\sum_{k}A_{b,i,k}B_{b,k,j}$. FTuner average is 21\% better than the Vendor, 43\% better than Roller, and 5\% better than HAOTuner.}
    \label{fig:bmm}
\end{figure}

\subsection{Dynamic Shape Tensor Program Performance}
\label{section6.1}


\textbf{Dense operator.}
Fig. \ref{fig:denseV100} shows the execution time of the Dense operator for 128 shapes on the V100. T represents the sequence length, which corresponds to the dynamic axis. Dense has one dynamic axis, while BatchMatmul has two. FTuner supports up to four dynamic axes. FTuner achieves comparable performance (within 10\%) on 44\% shapes. FTuner is up to 64\% better than Roller \cite{roller}, 23\% on average. Only 4\% of the shapes are lower than Roller because Roller can achieve strict alignment with hardware in these shapes. FTuner outperforms the HAOTuner \cite{hao} in 59\% of the shapes.

FTuner performs worse than HAOTuner in two intervals. Shapes 8-16 and 52-64 launch fewer blocks, resulting in lower SM utilization and a longer execution time. Users can customize the penalty coefficients in SIA based on scenarios to improve performance on high-frequency shapes. This issue does not exist on the K80 since a single operator cannot fully utilize the entire SM on an ample resources device (V100). We achieved relatively good results on the A100 as shown in Fig. \ref{fig:denseA100}. FTuner can achieve up to 66\% speedup over Roller when T=5 and can achieve 33\% speedup on average. FTuner outperforms the HAOTuner on 98\% of the shapes and can get up to 42\% speedup over HAOTuner when T=17, 12\% on average.

Fig. \ref{fig:dense80} presents the evaluation of the K80 GPU. Compared with the Vendor, FTuner can achieve comparable performance (within 10\%) on 50\% of the shapes, outperforming with 33\% of them. FTuner outperforms Roller in performance across all shapes. FTuner is up to 70\% better than Roller, 28\% on average. FTuner outperforms HAOTuner on 88\% of the shapes. Due to page limitations, this manuscript does not include the results on the 3090 and 4900 GPUs. Overall, FTuner performs an average of 28\% better than Roller in the execution time of the Dense operator. Compared to HAOTuner, the performance fluctuates within 11\%.

\begin{figure}
    \centering
    \includegraphics[width=\linewidth]{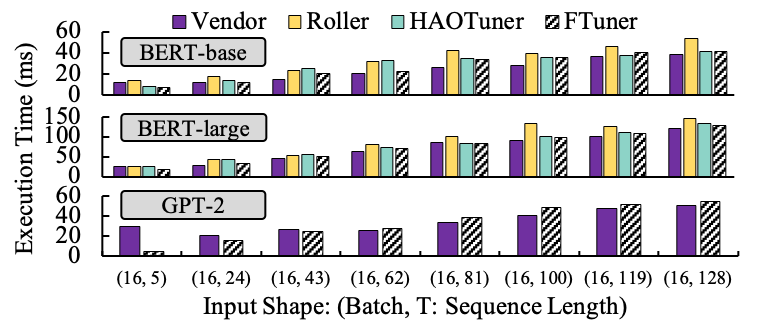}
    \caption{The execution time of the end-to-end models on V100 for 8 shapes. On BERT, FTuner differs from the Vendor's average by within 8\%. FTuner performs an average of 23\% better than Roller and 11\% better than HAOTuner. On GPT-2, FTuner is up to 5.6$\times$ better than Vendor when T=5, on average can achieve comparable performance. We could not measure HAOTuner and Roller on GPT-2.}
    \label{fig:dd}
\end{figure}

\textbf{BatchMatmul operator.} To maintain consistency with HAOTuner \cite{hao}, we sampled 8 shapes starting from 5 with a step of 19. As shown in Fig. \ref{fig:bmm}, FTuner can bring a 21\% performance improvement compared to Vendor. FTuner is up to 80\% better than Roller \cite{roller}, 43\% on average. FTuner is up to 51\% better than HAOTuner, 5\% on average. BatchMatmul has one extra axis in the input shape compared to Dense, with three dimensions along the space axis. Therefore, for HAOTuner, more dimensions expand the search space, making it hard for limited trials to find the best result. As the shape increases, the vendor can use larger loop tiles than FTuner and HAOTuner. When the shape is small, manually optimized tiles are difficult to cover. Conversely, as the shape increases, the vendor's performance becomes better. Due to the current version of cuBLAS being incompatible with the early Kepler architecture, the performance of the Vendor on K80 is noticeably inferior to others.



\textbf{End-to-end model.} We evaluated the execution time of three different end-to-end networks on the V100. As shown in Fig. \ref{fig:dd}, on BERT \cite{bert}, FTuner achieves performance comparable to Vendor (within 8\% on average). FTuner is up to 34\% better than HAOTuner \cite{hao}, 11\% on average. FTuner is up to 44\% better than Roller, 23\% on average. Since HAOTuner's auto-scheduling does not support the Dense+biasadd (a fused operator), it would interrupt tuning. Similarly, Roller does not support GPT-2, so we could not measure them. Due to exceeding the maximum protobuf size limit (2GB) during its graph loading phase, we could not verify the performance of BERT on TensorRT \cite{trt}.

\begin{figure}
    \centering
    \includegraphics[width=\linewidth]{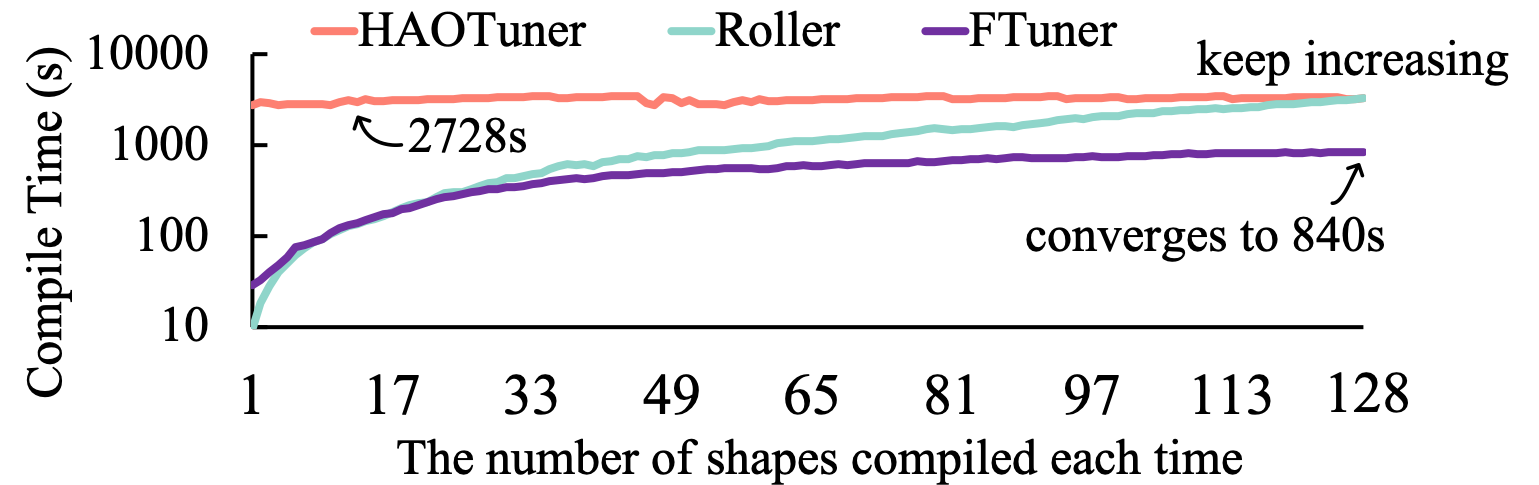}
    \caption{The compilation time. The number of shapes compiled each time increased from 1 to 128, using the Dense operator on V100. FTuner converges to 840s, Roller keeps increasing, and HAOTuner remains high.}
    \label{fig:op-time}
\end{figure}

\begin{figure}
    \centering
    \includegraphics[width=\linewidth]{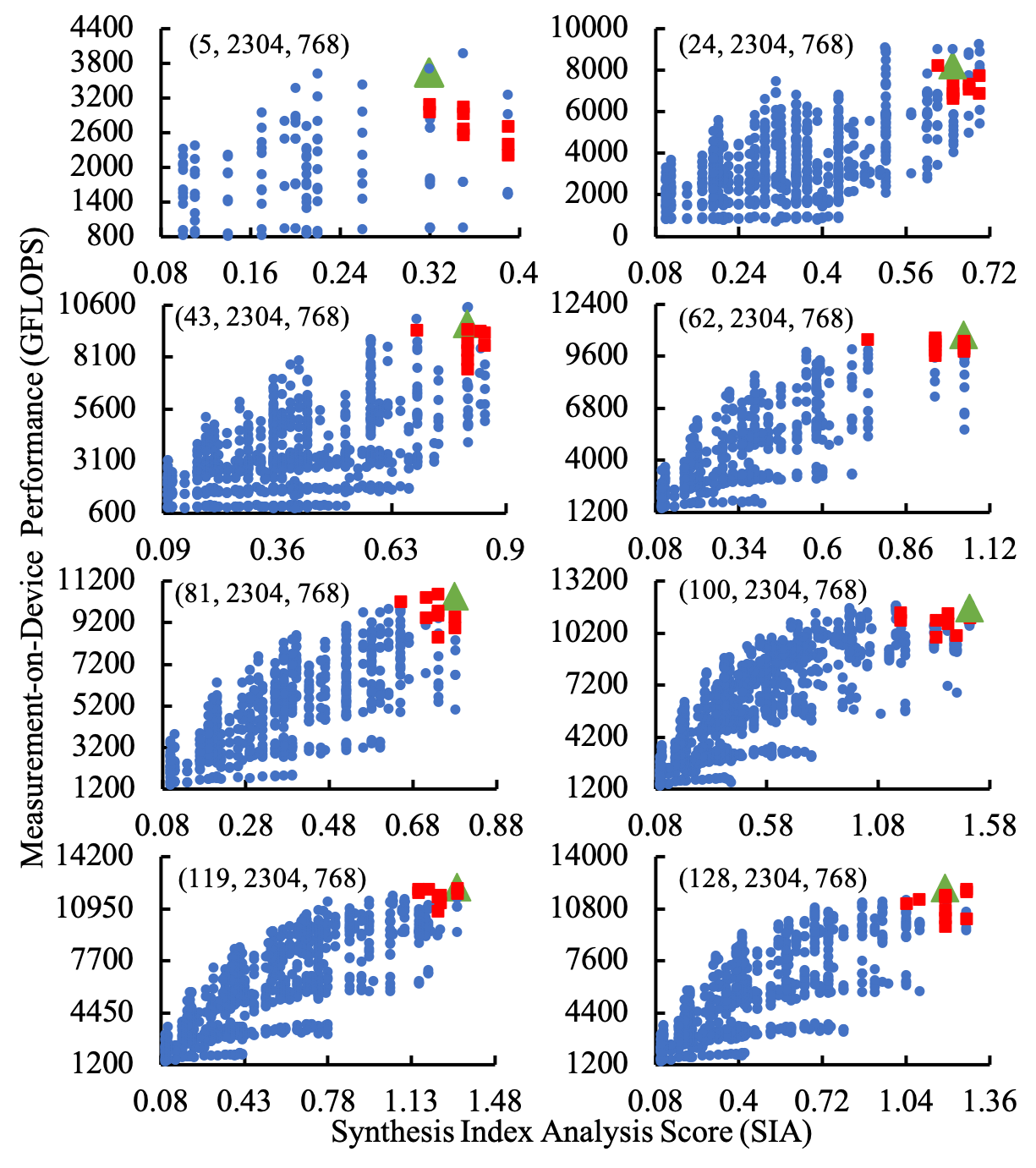}
    \caption{Evaluation of synthesis index analysis (SIA). Using the Dense operator on V100. \textbf{SIA can choose programs close to optimal performance at the upper right corner.} The dots represent programs generated by FTuner in uProg. The triangles and squares represent the Top1 and Top10 programs based on SIA scores, respectively.}
    \label{fig:scatter}
\end{figure}

\begin{table}
    \centering
    \renewcommand\arraystretch{1.1}
    \caption{Compilation time of single operators and end-to-end models for 8 input shapes on V100 GPU.}
    \resizebox{\linewidth}{!}{
        \begin{tabular}{c|ccccc}
        \hline
        &Ansor &HAOTuner &Roller &FTuner \\
        \hline
        Dense &6872s  &1104s  &163s  &149s \\
        \hline
        BatchMatmul &13549s  &2117s  &272s  &193s \\
        \hline
        BERT-base &69518s  &10862s  &2528s  &474s \\
        \hline
        BERT-large &103878s  &16231s  &3792s  &645s \\
        \hline
        GPT-2 &Time-out  &$-$  &$-$  &1614s \\
        \hline
    \end{tabular}
    }
    \label{tab:model-time}
\end{table}

\subsection{Compilation Time}
\label{section6.2}
We compare the compilation time of FTuner with HAOTuner \cite{hao} and Roller \cite{roller}. To comprehensively evaluate the results of FTuner and to maintain consistency with HAOTuner, we evaluate the total optimization time of the compiler for 8 different shapes. As shown in Fig. \ref{fig:op-time}, the compilation time of the Dense across 1 to 128 shapes. For one shape, the compilation time of FTuner is about 37.84s, Roller is 17.26s, and HAOTuner is 2728.86s. As the shapes increase, FTuner's compilation time gradually converges to 840s, while Roller continues to expand, and HAOTuner remains unchanged.


As shown in Tab. \ref{tab:model-time}, FTuner reduces two orders of magnitude for end-to-end models compared to Ansor \cite{ansor} and HAOTuner \cite{hao}. Compared to Roller \cite{roller}, FTuner reduces compilation time by one order of magnitude. In BERT-base, Ansor \cite{ansor} takes about 19.3 hours because Ansor requires recompilation whenever the input shape changes. HAOTuner \cite{hao} still needs 3 hours. The Roller is relatively faster and also takes 1.8 hours. FTuner completes the compilation process in only 474s, 645s for BERT-large, and 1614s for GPT-2. In the real world, the shapes of input tensors are arbitrary and far exceed 8. Therefore, the more shapes there are, the more obvious the advantage of FTuner in compilation time.

\subsection{Program Evaluation versus Measurement}
\label{section6.3}
To verify the accuracy of synthesis index analysis, we conducted measurement-on-device on all programs generated by FTuner. As shown in Fig. \ref{fig:scatter}, we used the SIA score as the horizontal axis and actual program throughput on the vertical axis. The dots represent programs in the program pool. The Top10 SIA are shown as squares, and the triangle represents the Top1. As the throughput increases, the SIA score generally shows an upward trend. However, there are inaccuracies in evaluations when the shape is small. Because SIA may not accurately select the highest performance and this evaluation method is simplistic, it cannot fully replace on-device measurement. However, this approach excels in the vast search space of dynamic shapes, allowing us to control performance evaluation errors within 10\%.

\begin{figure}
    \centering
    \includegraphics[width=\linewidth]{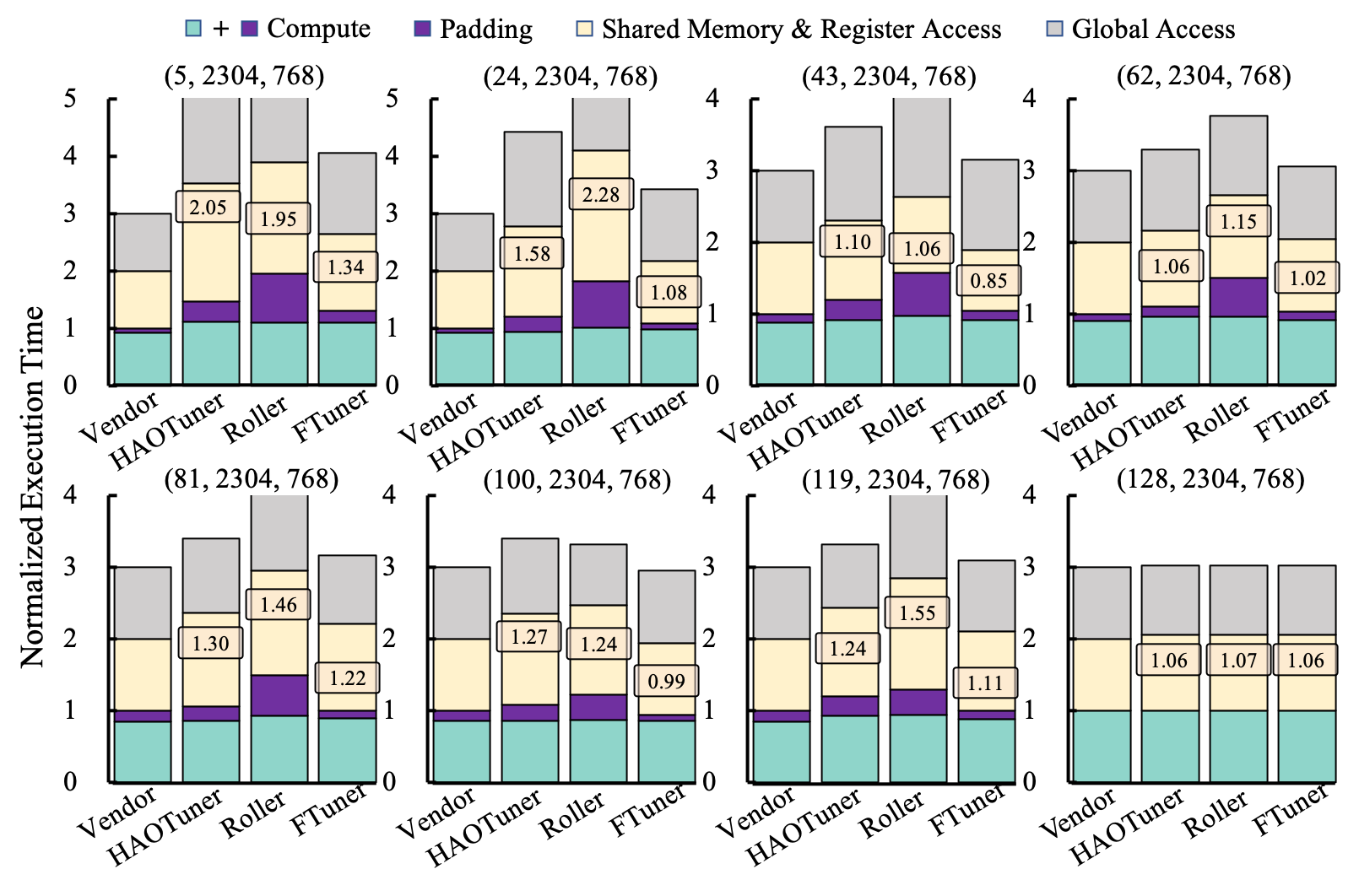}
    \caption{The proportion of padding in computation. Dense operator on V100 with 8 shapes. The vendor normalizes padding and access time. We can reduce the proportion of padding in computations to within 15\%. Compared to Roller, we have reduced it by up to 7.12$\times$, with an average reduction of 4.31$\times$.}
    \label{fig:ablation}
\end{figure}

\subsection{Padding and Memory Access Analysis}
\label{section6.4}
In this subsection, we analyze the proportion of padding in the computation time. As shown in Fig. \ref{fig:ablation}, we break down the computation time into effective computation (blue blocks) and ineffective computation, which is represented by the padding portion (mesh blocks). Compared to Roller \cite{roller}, FTuner can reduce padding by an average of 4.31$\times$. For shared memory access time, FTuner only exceeds Vendor by 8\%. Compared to HAOTuner \cite{hao} and Roller \cite{roller}, it reduces by 25\% and 39\%, respectively. As for global memory access time, FTuner exceeds Vendor by 14\%. Compared to HAOTuner and Roller, it reduces by 11\% and 13\%, respectively.




We also evaluated the SM utilization using the NVIDIA NCU Tool, as shown in Tab. \ref{tab:occ}. FTuner is up to 15\% better than Vendor, 4\% on average. FTuner is up to 1.2$\times$ better than HAOTuner, 35\% on average. FTuner is up to 37\% better than Roller, 15\% on average. This significant improvement in SM utilization is attributed to the multi-axis analysis of input tensors conducted in \S \ref{section4.4}.

\begin{table}
    \centering
    \renewcommand\arraystretch{1.1}
    \caption{The SM utilization on Dense operator.}
    \resizebox{\linewidth}{!}{
        \begin{tabular}{c|ccccc}
        \hline
        Input Shapes &Vendor &HAOTuner &Roller &FTuner \\
        \hline
        (5, 2304, 768) &63.91  &31.4  &75.84  &69.36 \\
        \hline
        (24, 2304, 768) &78.21  &51.87  &65.78  &74.69 \\
        \hline
        (43, 2304, 768) &89.39  &71.15  &60.12  &82.28 \\
        \hline
        (62, 2304, 768) &86.66  &66.4  &77.27  &75.14 \\
        \hline
        (81, 2304, 768) &72.71   &67.05  &67.51  &81.49 \\
        \hline
        (100, 2304, 768) &70.61   &67.89  &81.25  &83.13 \\
        \hline
        (119, 2304, 768) &73.37  &65.17  &58.48  &79.23 \\
        \hline
        (128, 2304, 768) &72.74  &67.03  &69.42  &83.3 \\
        \hline
    \end{tabular}
    }
    \label{tab:occ}
\end{table}

\section{Related Work}
\textbf{Deep learning compilers.} Well-featured and widely used deep learning compilers, such as XLA \cite{xla}, TC \cite{tc}, TVM \cite{tvm}, MLIR \cite{mlir}, Ansor \cite{ansor}, Alpa \cite{alpa}, and Roller \cite{roller}, FlexTensor \cite{flex}, Heron \cite{heron}, Meta-schedule \cite{meta}, AMOS \cite{amos}, Hidet \cite{hidet}, EINNET \cite{einnet}, can achieve excellent tuning results for tensor programs. However, when it comes to dynamic shape, the tuning process may not be implemented efficiently due to design limitations. BlazerML \cite{blazerml} and \cite{combin} Using multiple kernels on the CPU for Conv2D.


\textbf{Compilers for dynamic shape workloads.} Selective tuning \cite{selective-tuning} groups workloads into clusters based on similarity ratios and applies exact schedules to different shapes. Nimble \cite{nimble} generates a schedule for the largest shape and applies it to all shapes by loop tiling. DLight \cite{dlight} is a site package under development for dynamic shapes in the TVM community. DISC \cite{disc} builds a compiler based on MLIR \cite{mlir} but increases the memory footprint by introducing an offline compilation warm-up. CoRa \cite{cora} introduces a new set of scheduling primitives for dynamic shapes and uses uninterpreted functions \cite{cora-method} to symbolically represent variable loop bounds and scheduling operations. DISC \cite{disc} and CoRa \cite{cora} do not support auto-tuning. MikPoly \cite{mikpoly} constructs optimized programs for any shape on the fly by dividing the generation of micro-kernels into online and offline stages, but it still requires training in a cost model. Therefore, FTuner and MikPoly differ significantly in tuning time, and because MikPoly does not have open-source code, we cannot compare them.


\textbf{Learning-based cost model.}
TenSet\cite{tenset}, Moses\cite{moses}, TLP \cite{tlp} propose MLP-based pre-trained cost models for tensor compilers, but they require a public multi-platform dataset. In addition to supervised learning approaches, there are reinforcement learning approaches for compiler optimizations, such as Chameleon\cite{chameleon} and Neurovectorizer\cite{neurovectorizer}. TensorIR \cite{tir} is an intermediate representation that supports dynamic shapes. However, extensive measurement-on-device is required to obtain training data for the cost model.

\section{Conclusion}
\label{section8}
This paper introduces a fast dynamic shape auto-tuning compiler FTuner, which improves performance and reduces tuning time. FTuner utilizes an abstract computing unit called uKernel to compose dynamic shape tensors. It improves program performance by minimizing padding and provides a synthesis index analysis without a training cost model. FTuner outperforms other deep learning compilers in performance and tuning time. FTuner is portable, so we have adapted it to the Ascend architecture. When deploying to different hardware architectures, only the hardware description script needs to be provided, including memory bandwidth at different hierarchies, hardware core frequency, and number of hardware cores. In the future, we plan to refine CodeGen to support a broader backend. Additionally, as FTuner currently relies on numerous hyperparameters, we will explore an adaptive approach.






\bibliographystyle{unsrt}
\bibliography{references}

\end{document}